\documentclass[conference]{IEEEtran}

\usepackage{algorithm}
\usepackage{algpseudocode}
\usepackage{amsmath}
\usepackage{float}
\usepackage{flushend}
\usepackage{graphicx}
\usepackage{lettrine}
\usepackage{multirow}
\usepackage{subfigure}
\usepackage{threeparttable}

\begin{document}

\title{Real-Coded Chemical Reaction Optimization with Different Perturbation Functions}

\author{James J.Q. Yu, \textit{Student Member, IEEE}\\
       Department of Electrical and\\
       Electronic Engineering\\
       The University of Hong Kong\\
       Email: jqyu@eee.hku.hk\\
\and Albert Y.S. Lam, \textit{Member, IEEE}\\
       Department of Electrical Engineering\\
       and Computer Sciences\\
       University of California, Berkeley\\
       Email: albertlam@ieee.org\\
\and Victor O.K. Li, \textit{Fellow, IEEE}\\
       Department of Electrical and\\
       Electronic Engineering\\
       The University of Hong Kong\\
       Email: vli@eee.hku.hk\\
}

\maketitle
\pagestyle{empty}

\begin{abstract}
Chemical Reaction Optimization (CRO) is a powerful metaheuristic which mimics the interactions of molecules in chemical reactions to search for the global optimum. The perturbation function greatly influences the performance of CRO on solving different continuous problems. In this paper, we study four different probability distributions, namely, the Gaussian distribution, the Cauchy distribution, the exponential distribution, and a modified Rayleigh distribution, for the perturbation function of CRO. Different distributions have different impacts on the solutions. The distributions are tested by a set of well-known benchmark functions and simulation results show that problems with different characteristics have different preference on the distribution function. Our study gives guidelines to design CRO for different types of optimization problems.
\end{abstract}


\section{Introduction}

\lettrine[lines=2]{C}{hemical} Reaction Optimization (CRO) \cite{AYSLam2010} is a simple and powerful metaheuristic optimization method which mimics the interactions of molecules in chemical reactions to search for the global optimum. CRO was designed as an optimization framework and it was initially targeted to solve discrete optimization problems \cite{JXu2011}\cite{AYSLam2010b}\cite{AYSLam2012}. It has been applied to solve many practical problems, e.g. population transition problem in peer-to-peer streaming \cite{AYSLam2010c}, network coding optimization problem \cite{BPan2011}, etc. Lam \textit{et al.} then proposed a variant of CRO in 2011, named Real-Coded Chemical Reaction Optimization (RCCRO) \cite{AYSLam2011}, to solve continuous optimization problems. RCCRO utilizes the Gaussian distribution function as the perturbation function and some real-coded-based mechanisms were designed to implement RCCRO. RCCRO has been shown to be efficient in solving continuous optimization problems \cite{AYSLam2011}\cite{JJQYu2011}.

There are four major operations (i.e., elementary reactions) in CRO: on-wall ineffective collision, decomposition, inter-molecular ineffective collision, and synthesis. In CRO, on-wall ineffective collision and inter-molecular ineffective collision correspond to local search, while decomposition and synthesis correspond to remote search. In the conventional RCCRO \cite{AYSLam2011}, the Gaussian distribution is deployed in the neighborhood search operator in all reactions except synthesis. In these elementary reactions, with the Gaussian distribution, the molecules are perturbed via the Gaussian mutation and the energy state of the molecules are checked to decide whether the reaction shall be accepted or rejected. The Gaussian perturbation in RCCRO is accomplished by adding a zero-mean Gaussian-random number to the existing molecular structures (i.e., solutions) to generate new solutions in the neighborhoods.

For other optimization methods like Evolutionary Programming (EP) and Particle Swarm Optimization (PSO), researchers have also proposed other perturbation functions other than the Gaussian distribution. Mutation operations in EP based on the Cauchy distribution \cite{XYao1996} and the Levy distribution \cite{CLee2004} have been propose. Krohling \textit{et al.} has made contributions in integrating PSO with the exponential \cite{RAKrohling2006}, the Gaussian, and the Cauchy distribution \cite{RAKrohling2009}. However, the impacts of different perturbation distributions on the performance of RCCRO need further study. Such research is necessary for a better understanding of the performance of RCCRO.

In this paper, we propose to apply four different probability distribution functions as the perturbation function for RCCRO, namely, the Gaussian distribution, the Cauchy distribution, the exponential distribution, and a modified Rayleigh distribution. These distributions have drawn much attention from different research communities \cite{CLee2004}\cite{BMandelbrot1982}. These four distributions are tested on a suite of well-known benchmark functions classified into three categories of optimization functions. The simulation results demonstrate that different categories of functions have different preferences on the perturbation distribution function.

The rest of the paper is organized as follows: Section II presents an overview of CRO. In Section III, the four tested distribution functions are described. Section IV provides the simulation results. The paper is concluded in Section V with suggestions for further research.

\section{Chemical Reaction Optimization}
In this section we will first introduce the manipulated agents of CRO, i.e., molecules. Then the definitions of elementary reactions are presented. The section ends with the overall algorithm of RCCRO (or simply referred to CRO hereafter).

\subsection{Molecules}
CRO is a kind of population-based metaheuristics \cite{AYSLam2010}. There is a population of molecules in a container, with an energy buffer attached. Each molecule is characterized by its molecular structure ($\omega$), potential energy (PE), kinetic energy (KE), and some other attributes. $\omega$ stands for a feasible solution of the optimization problem, corresponding to this molecule. PE represents the the objective function value of $\omega$ while KE represents the tolerance of the molecule to hold a worse solution with larger objective function value than the existing one. Other attributes can be used to control the flow of CRO to ensure CRO meet the characteristics of the optimization problem. Users can add, change, or remove the optimal attributes to build different versions of CRO.

\subsection{Elementary Reactions}
In CRO, changes to the molecules are made through different elementary reactions. There are four kinds of elementary reactions, namely, on-wall ineffective collision, decomposition, inter-molecular ineffective collision, and synthesis. The former two reactions belong to the uni-molecular reactions and the latter two are classified as class of inter-molecular reactions. These elementary reactions change the molecular structures of the molecules and accept molecules with new structures according to the conservation of energy. CRO makes use of these changes to explore the solution space and locates the global optimum \cite{AYSLam2010}. We basically follow \cite{AYSLam2011} to define the elementary reactions and they are described briefly as follows:

\subsubsection{On-wall Ineffective Collision}
An on-wall ineffective collision happens when a molecule collides with a wall of the container and then bounces away. This reaction has one molecule as input and returns another modified molecule. This reaction is mainly used for performing local search, thus the changes made on the molecular structure shall be small. We commonly generate a neighborhood structure $\omega^{\prime}$ from $\omega$. If we define the neighbor function as
\begin{equation}
\textit{neighbor}(\omega)=\omega+\epsilon,
\end{equation}
where $\epsilon$ is called the perturbation factor and is generated from a pre-defined probability distribution function, which will be elaborated in Section IV-E. We obtain a new solution $\omega^\prime$ by
\begin{equation}
\omega^{\prime}=\textit{neighbor}(\omega)
\end{equation}
and its new PE is fiven by
\begin{equation}
\textit{PE}_{\omega^{\prime}}=f(\omega^\prime),
\end{equation}
where $f$ is the objective function.

During this procedure, a part of the KE held by the molecule will be transfered to the system energy buffer (\textit{EnBuff}). We have a parameter \textit{LossRate} to control this energy loss process. The new KE of the molecule will be updated according to
\begin{equation}
\textit{KE}_{\omega^{\prime}}=(\textit{PE}_{\omega}-\textit{PE}_{\omega^{\prime}}+\textit{KE}_{\omega})\times t,
\end{equation}
where $t\in[LossRate,1]$ is a randomly generated number.

\subsubsection{Decomposition}
A decomposition happens when a molecule collides with a wall of the container and breaks into two molecules. This reaction is mainly used for jumping out of the local minimums, and thus the changes made on the molecular structures are larger than an on-wall ineffective collision. During this procedure, there is no KE transferred to \textit{EnBuff}, but the energy conservation law shall hold \cite{AYSLam2011}.

\subsubsection{Inter-molecular Ineffective Collision}
An inter-molecular ineffective collision happens when two molecules collide with each other and then separate. The purpose and characteristics of this reaction is similar with an on-wall ineffective collision. In general, we perform two neighbor searches on the two molecular structures $\omega_{1}$ and $\omega_{2}$, i.e.,
\begin{equation}
\omega_{1}^{\prime}=neighbor(\omega_{1})\text{ and }\omega_{2}^{\prime}=neighbor(\omega_{2}).
\end{equation}

\subsubsection{Synthesis}
A synthesis happens when two molecules collide and merge into one. Generally, the change is severe and can help the molecule jump out of local minimums. Its objective is to maintain the population diversity.

In this operator, the new molecule is derived from the two given original molecules and each element of the solution is equally likely to be selected from each of the original molecules at the same position.

\subsection{The Overall Algorithm}
CRO operates in a closed container with an initial population of randomly generated molecules. The algorithm contains three phases: initialization, iterations, and finalization. When CRO starts, the molecules as well as some system parameters, such as \textit{EnBuff}, \textit{CollRate}, \textit{LossRate}, \textit{DecThres}, and \textit{SynThres} \cite{AYSLam2011}\cite{JJQYu2012} are set. In each iteration, the system will first randomly select one reaction type according to some criteria. The system will then use this decision to randomly select one or multiple molecules from the existing ones in the container depending on whether it is a uni-molecular or an inter-molecular reaction and check its/their energy state(s). If the decomposition criterion (for uni-molecular collision) or the synthesis criterion (for inter-molecular collision) described in \cite{AYSLam2010} is satisfied, the corresponding reaction takes place. Otherwise an on-wall ineffective collision or an inter-molecular ineffective collision will take place. At the end of each iteration is the energy check. The newly generated or transformed molecule(s) have their objective functions evaluated and compared with the original molecules. If the new value(s) can satisfy the energy conservation conditions \cite{AYSLam2010}, the new molecules are accepted and put into the container while the original molecules are discarded. Otherwise, the new molecules are discarded and this indicates a failed reaction. After the pre-defined certain number of function evaluations is reached or one of some specific stopping criteria is met, the algorithm proceeds to finalization and outputs the best-so-far global optimum. Since in this problem we focus on studying the performance of CRO with different probability distributions used in the neighborhood search operator and decomposition stated in Section II-B, interested readers can refer to \cite{AYSLam2010} and \cite{AYSLam2011} for detailed description of the algorithm and its pseudocode.

\section{Probability Distribution}
In this paper, we consider four different types of probability distributions, namely, the Gaussian distribution, the Cauchy distribution, the exponential distribution, and the Rayleigh distribution. In this section, we will first introduce the four distributions. Then the integration of CRO with these distributions is presented.

\subsection{Gaussian Distribution}
The Gaussian distribution has a bell-shaped probability density function (PDF) and is also known as the normal distribution or informally the bell curve \cite{GCasella2001}. Its density function is given as:

\begin{equation}
f(x;\mu,\sigma)=\frac{1}{\sqrt{2\pi\sigma^{2}}}e^{-\frac{(x-\mu)^{2}}{2\sigma^{2}}},
\end{equation}
where $\mu$ is the expectation and $\sigma^{2}$ is the variance. The plots of the PDF with some different $\mu$ and $\sigma^{2}$ are given in Fig. 1.

\begin{figure}
	\centering
	\includegraphics[width=0.5\textwidth]{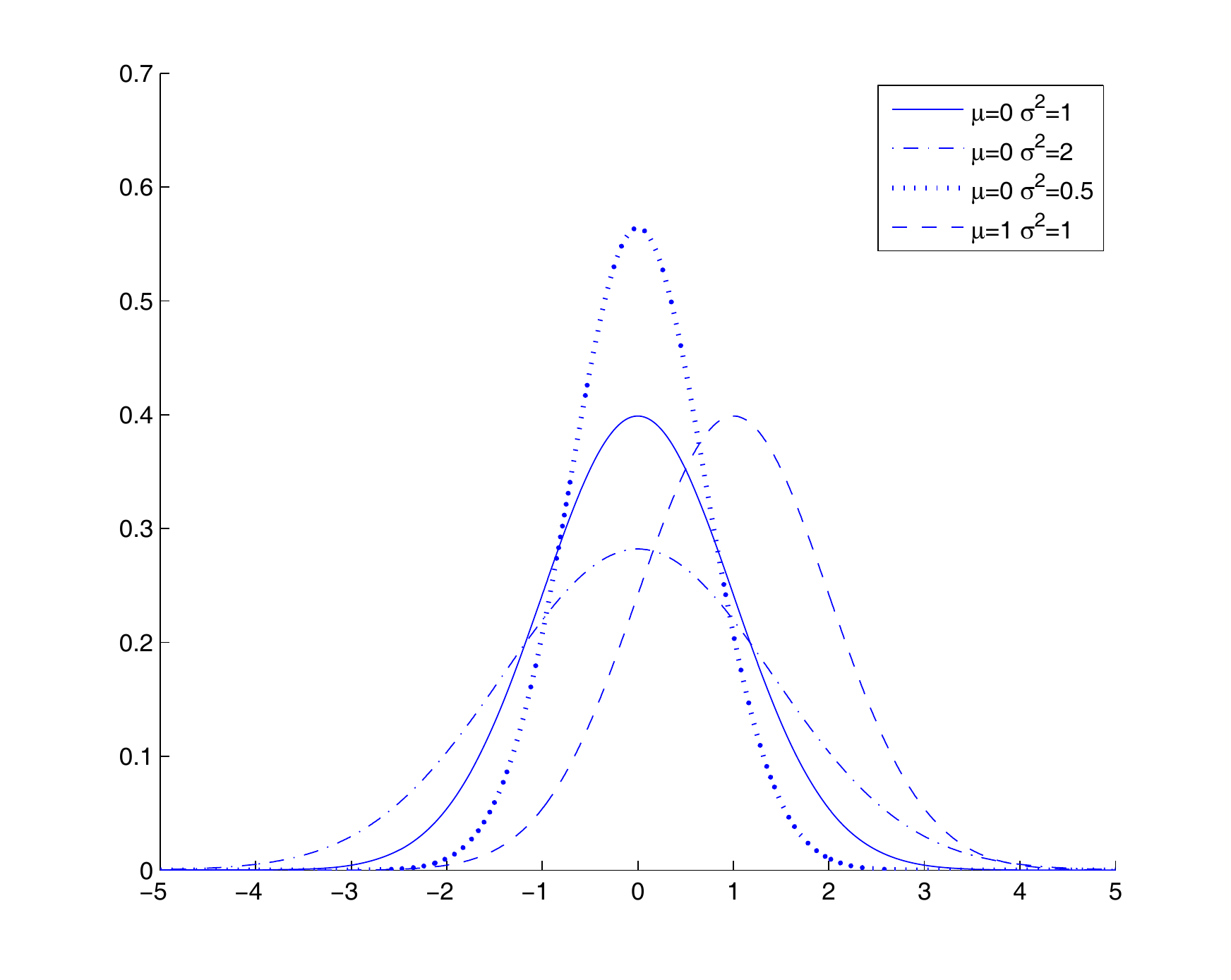}
	\caption{PDF of Gaussian Distribution}
\end{figure}

We can control the shape by modifying the value of $\sigma^{2}$. A larger $\sigma^{2}$ will result in a flatter bell. 

\subsection{Cauchy Distribution}
The Cauchy distribution shares a similar bell shape with the Gaussian distribution and it has important applications to physics. We can utilize its characteristics to perform detailed local search. Its density function is given as:

\begin{equation}
f(x;x_{0},\gamma)=\frac{1}{\pi}[\frac{\gamma}{(x-x_{0})^{2}+\gamma^{2}}].
\end{equation}

In this PDF, $x_{0}$ is the location parameter, or mean of the distribution. It is similar to $\mu$ in the Gaussian distribution. $\gamma$ is the scale parameter which can specify the half-width at half-maximum (HWHM). The plots of the Cauchy distribution with different $x_{0}$ and $\gamma$ are given in Fig. 2.

\begin{figure}
	\centering
	\includegraphics[width=0.5\textwidth]{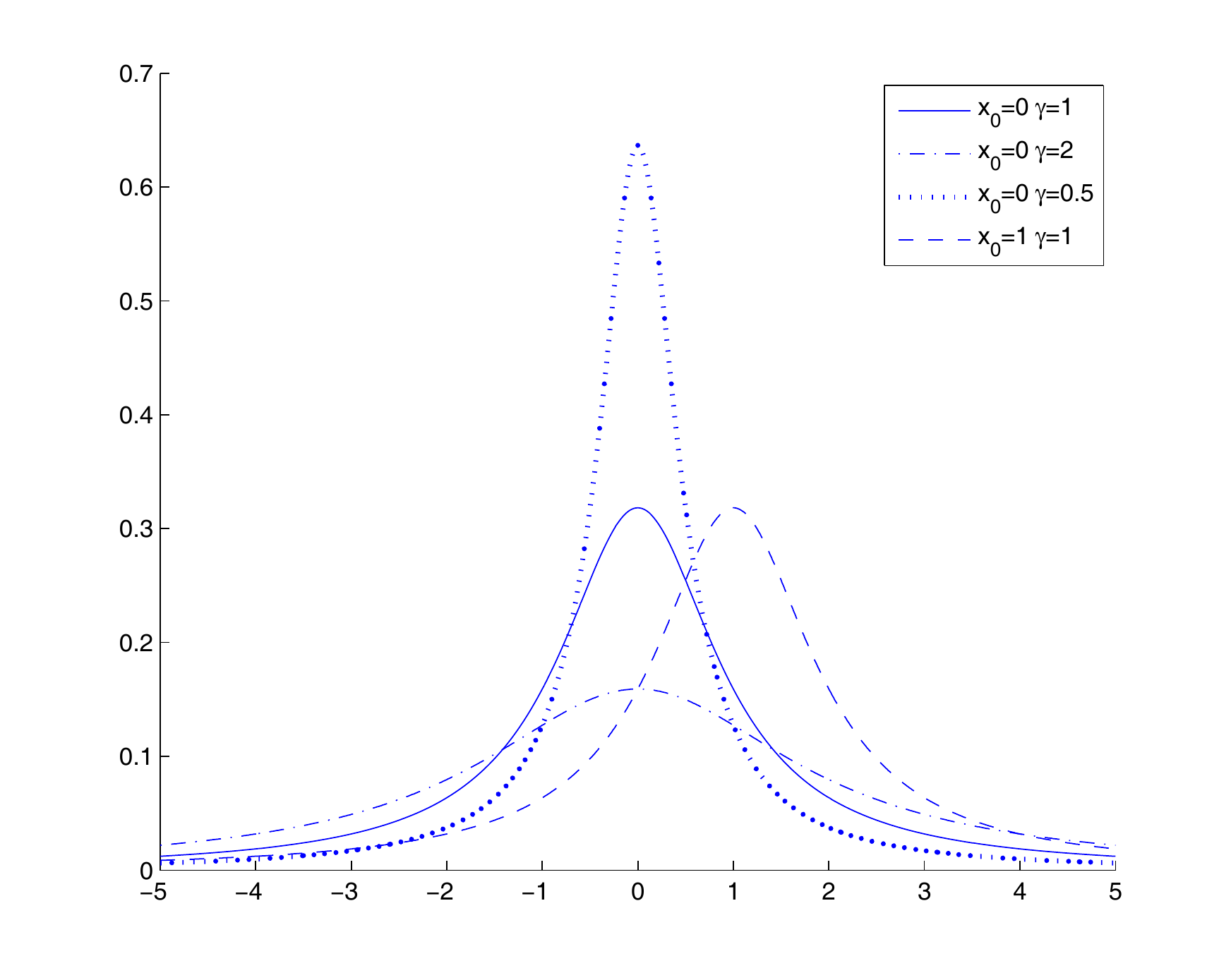}
	\caption{PDF of Cauchy Distribution}
\end{figure}

\subsection{Exponential Distribution}
The exponential distribution describes the time between events in a Poisson process. Different from the previous two distributions, the exponential distribution does not have a bell-shape. Its density function is given as:

\begin{equation}
f(x;x_{0},\gamma)=\left\{
\begin{aligned}
& \gamma e^{-\gamma (x-x_{0})}, & x\geq0 \\
& 0, & x<0 
\end{aligned}
\right.
\end{equation}

The parameter $x_{0}$ defines the starting point of the distribution and the parameter $\gamma$ defines the steepness of the PDF curve. The plots of the distribution with different $x_{0}$ and $\gamma$ are given in Fig. 3.

\begin{figure}
	\centering
	\includegraphics[width=0.5\textwidth]{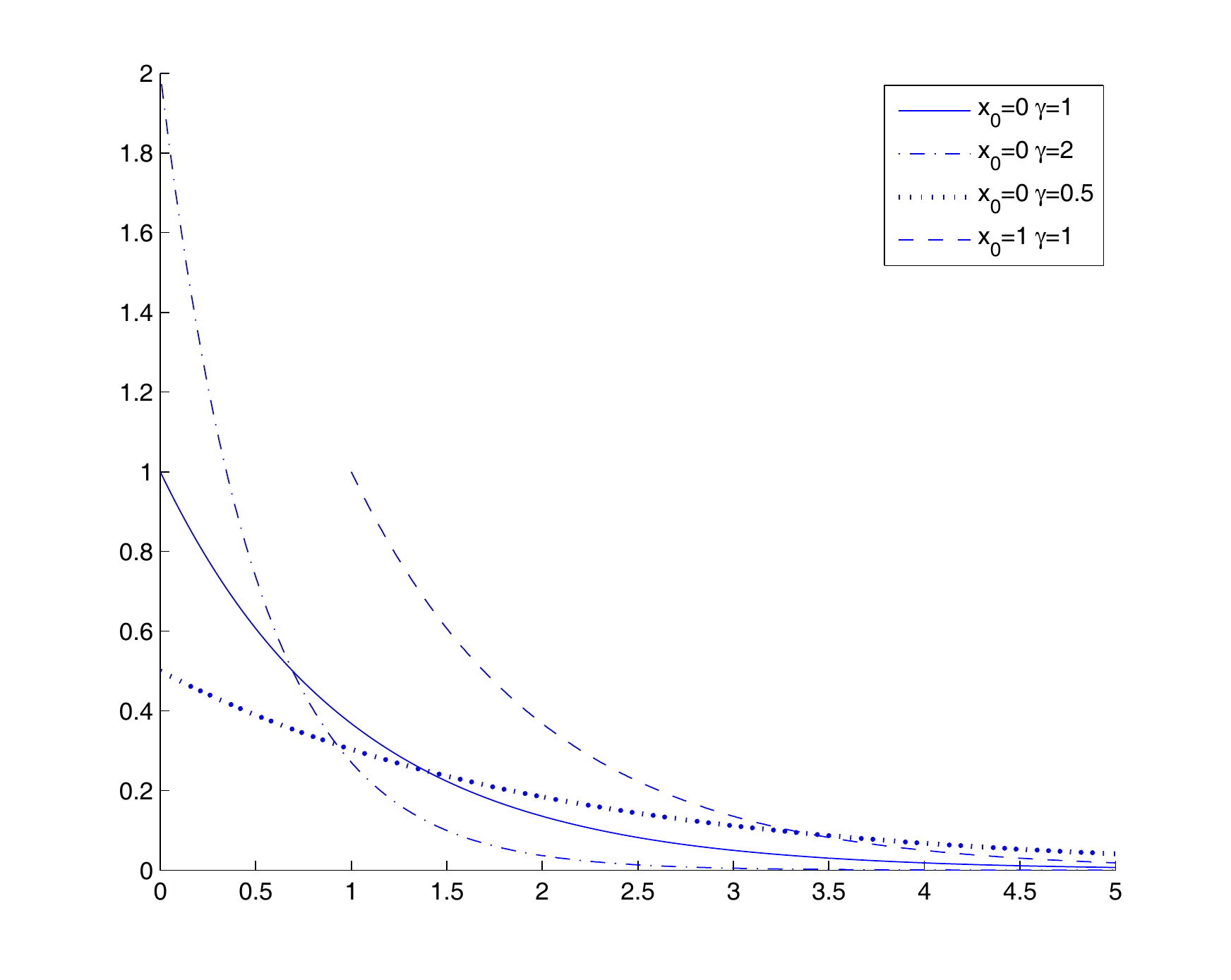}
	\caption{PDF of Exponential Distribution}
\end{figure}

Note that the normal exponential distribution only has a positive side, which does not satisfy our requirement as a perturbation function. Similar to \cite{RAKrohling2006}, we mirror this curve to the negative side of the x-axis to make this distribution a continuous function on the range $(-\infty,\infty)$.

\subsection{Rayleigh Distribution}
The Rayleigh PDF is given below:

\begin{equation}
f(x;\sigma^{2})=\frac{x}{\sigma^{2}}e^{-\frac{x^{2}}{2\sigma^{2}}}, x\geq0.
\end{equation}

As with  the Gaussian distribution, the only parameter, $\sigma^{2}$ controls the flatness of the shape. A larger $\sigma^{2}$ results in a flatter PDF. Moreover, the highest point of the curve occurs at $x=\sigma$. The plots of the Rayleigh distribution with different $\sigma^{2}$ are given in Fig. 4.

\begin{figure}
	\centering
	\includegraphics[width=0.5\textwidth]{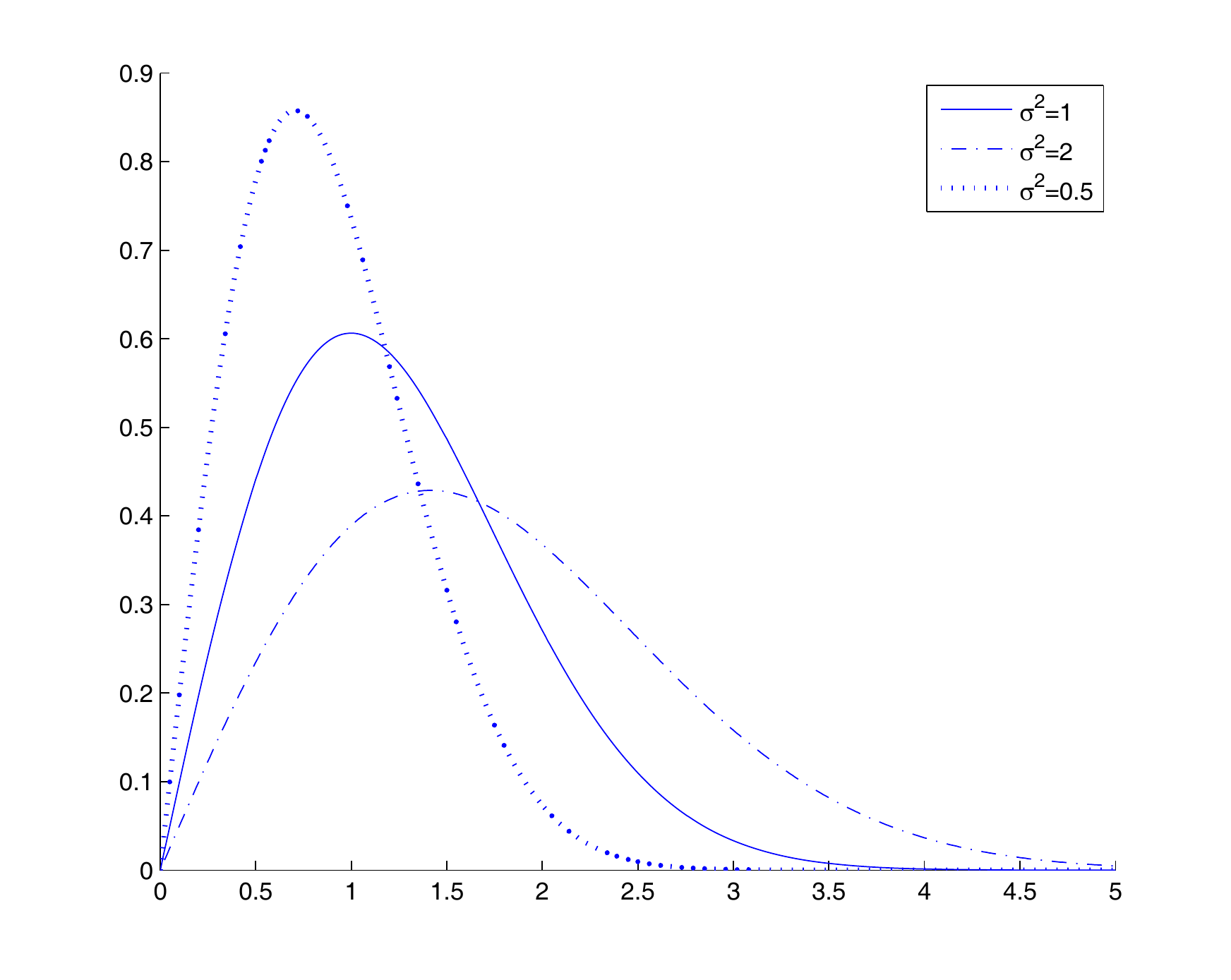}
	\caption{PDF of Rayleigh Distribution}
\end{figure}

Like the exponential distribution, the Rayleigh distribution is only continuous on $(0,\infty)$. So we introduce a new way to make the distribution function continuous on $(-\infty,\infty)$ and apply it to the perturbation function. The modified PDF is given below:

\begin{equation}
\begin{aligned}
f^{\prime}(x;\sigma^{2})=& [\frac{\sigma+x}{\sigma^{2}}e^{-\frac{(\sigma+x)^{2}}{2\sigma^{2}}}\times step(x,-\sigma)+ \\
& \frac{\sigma-x}{\sigma^{2}}e^{-\frac{(\sigma-x)^{2}}{2\sigma^{2}}}\times (1-step(x,\sigma))]/2
\end{aligned}
\end{equation}
where
\begin{equation}
step(x,\sigma)=\left\{
\begin{aligned}
& 1, & x\geq\sigma \\
& 0, & x<\sigma 
\end{aligned}
\right.
\end{equation}

This modified PDF is composed of two parts: We first shift the original PDF to the left by $\sigma$ units, which makes $x=\sigma$ as the y-axis. Then this new curve is copied and mirrored at the y-axis. The two curves are then summed and averaged. The plots of this modified PDF with different $\sigma^2$ values are given in Fig. 5.

\begin{figure}
	\centering
	\includegraphics[width=0.5\textwidth]{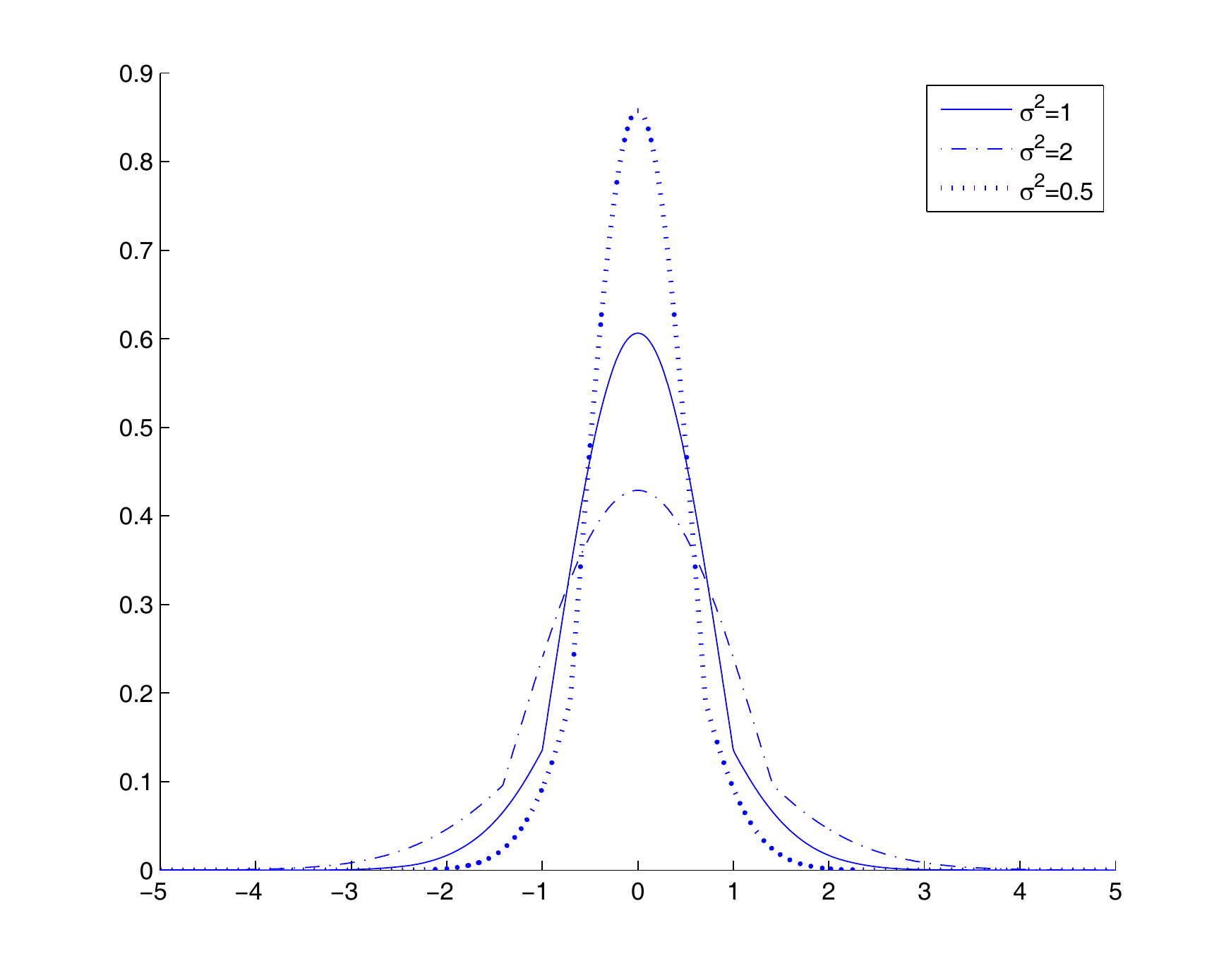}
	\caption{PDF of Modified Rayleigh Distribution}
\end{figure}

\subsection{Integration of Distributions with CRO}
The perturbation function mainly operates in the on-wall ineffective collision, inter-molecular ineffective collision, and decomposition \cite{AYSLam2011}. In the above three elementary reactions, new molecular structures are generated from the original structures. Although in synthesis there is also a new structure generated, the new structure is composed of different parts of the original structures. To generate new molecular structures, one can add a perturbation factor to one random element of the original structure for a small change (on-wall ineffective collision and inter-molecular ineffective collision), or add different perturbation factors to some of the elements for a dramatic change (decomposition) \cite{AYSLam2011}. The perturbation factors are generated from the perturbation function, which can be one of the pre-defined probability distributions discussed before. So different probability distribution functions shall generate perturbation factors with different characteristics, and this will result in different performances of the algorithm.

\section{Simulation Results}
\begin{table*}
	\caption{Parameter Settings}
	\small
	\begin{center}
	\begin{threeparttable}
		\begin{tabular}{l|l|l|l|l|l|l|l|l}
			\hline
			Category & PopSize & StepSize & EnBuff & IniKE & CollRate & LossRate & DecThres & SynThres \\ \hline
			I & 10 & 0.1 & $10^6$ & $10^3$ & 0.2 & 0.9 & $1.5\time10^5$ & 0 \\ 
			II & 20 & $1^*$ & $10^5$ & $10^7$ & 0.2 & 0.1 & $1.5\time10^5$ & 10 \\ 
			III & 100 & 0.5 & 0 & $10^3$ & 0.2 & 0.1 & 500 & 10 \\ \hline
		\end{tabular} 
		\begin{tablenotes}
		       \item[*] 300 for $f_8$ and 15 for $f_{11}$
     		\end{tablenotes}
	\end{threeparttable}
	\end{center}
\end{table*}

\subsection{Benchmark Functions}
In order to evaluate the performance of different perturbation distribution functions, we perform simulations on the standard benchmark functions used in \cite{XYao1999}. The benchmark functions are listed in \cite{AYSLam2011} with the dimension sizes, feasible solution spaces and the known global optimums. This set of functions has been widely used or partially used as metaheuristic performance evaluation benchmark functions \cite{AYSLam2011}\cite{XYao1999}\cite{HDong2007}.

This benchmark set can be divided into three categories according to their characteristics. The first group consists of unimodal functions, each of which has only one global optimum. So it is relatively easy to solve. The second group contains high-dimensional multimodal functions. Functions in this group have multiple local optimums and it is more likely for algorithms to get ``stuck" in the local optimums. The last group is a collection of low-dimensional multimodal functions. These functions have lower dimensions and fewer local optimums than the second group.

\subsection{Experiment Environment}
All the simulations are performed on the same computer with Intel Core i5-2400 @ 3.1GHz CPU and 4.00GB RAM. CRO and distribution functions are implemented using C++ and compiled with MingGW g++ compiler under the Windows 7 64bit environment.

\subsection{Parameter Selection}
Different parameter settings can affect the performance of CRO dramatically \cite{AYSLam2011}. A suitable combination of parameters, including \textit{PopSize}, \textit{StepSize}, \textit{EnBuff}, \textit{IniKE}, \textit{CollRate}, \textit{LossRate}, \textit{DecThres}, and \textit{SynThres} \cite{JJQYu2012}, may result in good simulation results. Since the combinations of parameters exist in an eight-dimensional space and are continuous, it is impractical to test all possible combinations. Instead, the parameters are tuned in an ad hoc manner. In our simulations presented in the next subsection, we adopt the three different parameter combinations for the three function categories presented and discussed in \cite{AYSLam2011}. The details of the combinations are listed in Table I and interested readers can refer to \cite{AYSLam2011} and \cite{JJQYu2012} for elaboration of the functionality of each parameter.

\subsection{Comparisons among Different Distributions}
In this paper we presented four different distribution functions as CRO's perturbation function, namely the Gaussian distribution, the Cauchy distribution, the exponential distribution, and a modified Rayleigh distribution. For simplicity, we denote the CRO algorithms with distributions as CRO\_G, CRO\_C, CRO\_E, and CRO\_R, respectively. We evaluate the performance of the four CRO variants  with the benchmark functions with the parameter combinations discussed in the previous subsection. We repeat the simulations for each function 100 times and the function evaluation limits for different functions are listed in Table II which are the same as those used in \cite{AYSLam2011}.

\begin{table}
	\caption{Function Evaluation Limits}
	\small
	\begin{center}
		\begin{tabular}{l|r|l|r}
			\hline
			Function & FELimit & Function & FELimit\\ \hline
			$f_{1}$ & 150 000 & $f_{13}$ & 150 000 \\
			$f_{2}$ & 150 000 & $f_{14}$ & 7 500 \\
			$f_{3}$ & 250 000 & $f_{15}$ & 250 000 \\
			$f_{4}$ & 150 000 & $f_{16}$ & 1 250 \\
			$f_{5}$ & 150 000 & $f_{17}$ & 5 000 \\
			$f_{6}$ & 150 000 & $f_{18}$ & 10 000 \\
			$f_{7}$ & 150 000 & $f_{19}$ & 4 000 \\
			$f_{8}$ & 150 000 & $f_{20}$ & 7 500 \\
			$f_{9}$ & 250 000 & $f_{21}$ & 10 000 \\
			$f_{10}$ & 150 000 & $f_{22}$ & 10 000 \\
			$f_{11}$ & 150 000 & $f_{23}$ & 10 000 \\ 
			$f_{12}$ & 150 000 & &\\ \hline
		\end{tabular} 
	\end{center}
\end{table}

The simulation results for the four algorithms on the 23 benchmark functions are presented in Table III, Table IV, and Table V, and the best results are bolded. Since the three categories of benchmark functions have different characteristics on the solution space, we will discuss them separately.

\begin{table*}
	\caption{Simulation Results and Comparisons for Category I}
	\small
	\begin{center}
		\begin{tabular}{l|llll|llll}
			\hline
			\multirow{2}*{Function} & \multicolumn{4}{c|}{CRO\_G} & \multicolumn{4}{c}{CRO\_C} \\ \cline{2-9}
			 & Mean & Std. Div. & Best & Rank & Mean & Std. Div. & Best & Rank\\ \hline
$f_1$ & \bf{2.8023E-06}&9.5462E-06&1.1729E-06&1&3.6134E-06&1.0076E-05&1.5009E-06&2\\
$f_2$ & \bf{5.2742E-03}&8.6993E-03&3.1498E-03&1&6.5312E-03&1.0253E-02&4.2567E-03&2\\
$f_3$ & \bf{4.0448E-07}&1.7591E-06&1.3886E-07&1&5.9970E-07&2.2993E-06&1.8863E-07&2\\
$f_4$ & 1.5898E+00&4.9408E+01&3.7482E-03&3&\bf{2.1603E-02}&3.2341E-01&3.4985E-03&1\\
$f_5$ & 7.9995E+01&1.3336E+03&1.7200E-02&4&\bf{4.9454E+01}&3.5096E+02&6.5425E-01&1\\
$f_6$ & \bf{0.0000E+00}&0.0000E+00&0.0000E+00&1&\bf{0.0000E+00}&0.0000E+0&0.0000E+00&1\\
$f_7$ & 1.0101E-02&6.0951E-02&2.6305E-03&2&\bf{9.0633E-03}&4.4733E-02&1.9008E-03&1\\ 
Avg. & & & &1.86& & & &\bf{1.43}\\ \hline
			\multirow{2}*{Function} & \multicolumn{4}{c|}{CRO\_E} & \multicolumn{4}{c}{CRO\_R} \\ \cline{2-9}
			 & Mean & Std. Div. & Best & Rank & Mean & Std. Div. & Best & Rank\\ \hline
$f_1$ &2.3209E-05&8.6462E-05&8.9015E-06&4&5.8125E-06&1.8736E-05&2.2348E-06&3\\
$f_2$ &1.6889E-02&2.6210E-02&1.0355E-02&4&7.7805E-03&1.0789E-02&4.7191E-03&3\\
$f_3$ &3.7256E-06&1.4036E-05&1.0769E-06&3&8.7830E-07&3.4124E-06&3.6419E-07&4\\
$f_4$ &4.6047E-02&6.6140E-01&1.0012E-02&2&2.4322E+00&7.1027E+01&4.4672E-03&4\\
$f_5$ &6.2050E+01&3.0865E+02&1.2512E+00&2&7.3270E+01&9.0927E+02&2.2321E+00&3\\
$f_6$ &\bf{0.0000E+00}&0.0000E+00&0.0000E+00&1&\bf{0.0000E+00}&0.0000E+0&0.0000E+00&1\\
$f_7$ &1.3602E-02&7.1639E-02&4.5453E-03&4&1.0531E-02&4.6002E-02&4.1566E-03&3\\
Avg. & & & &2.86& & & &3.00\\ \hline
		\end{tabular} 
	\end{center}
\end{table*}

\begin{table*}
	\caption{Simulation Results and Comparisons for Category II}
	\small
	\begin{center}
		\begin{tabular}{l|llll|llll}
			\hline
			\multirow{2}*{Function} & \multicolumn{4}{c|}{CRO\_G} & \multicolumn{4}{c}{CRO\_C} \\ \cline{2-9}
			 & Mean & Std. Div. & Best & Rank & Mean & Std. Div. & Best & Rank\\ \hline
$f_{8}$ & -6.6941E+03&7.0479E+03&-8.8963E+03&3&\bf{-1.1584E+04}&2.7988E+03&-1.2451E+04&1\\
$f_{9}$ & 7.3239E-04&5.1246E-03&2.8899E-04&3&9.6594E-04&2.9687E-03&4.8051E-04&4\\
$f_{10}$ & 2.3286E-03&7.6100E-03&1.6035E-03&3&3.0142E-03&1.1784E-02&1.6159E-03&4\\
$f_{11}$ & 6.3738E+00&4.4097E+01&3.9130E-01&3&\bf{6.9692E-02}&5.0228E-01&6.6663E-07&1\\
$f_{12}$ & 8.9318E-02&1.7332E+00&4.7978E-08&3&2.2222E-07&1.6140E-06&6.2396E-08&2\\
$f_{13}$ & 2.5472E-06&3.4365E-05&5.7443E-07&3&3.1977E-06&3.2292E-05&5.6288E-07&4\\ 
Avg. & & & &3.00& & & &2.67\\ \hline
			\multirow{2}*{Function} & \multicolumn{4}{c|}{CRO\_E} & \multicolumn{4}{c}{CRO\_R} \\ \cline{2-9}
			 & Mean & Std. Div. & Best & Rank & Mean & Std. Div. & Best & Rank\\ \hline
$f_{8}$ & -6.6816E+03&6.8112E+03&-8.6168E+03&4&-6.7458E+03&6.7677E+03&-8.1446E+03&2\\
$f_{9}$ & 4.0876E-04&1.5659E-03&1.2755E-04&2&\bf{3.4566E-04}&4.6215E-03&1.3204E-04&1\\
$f_{10}$ & 1.8143E-03&2.7791E-03&1.1840E-03&2&\bf{1.6536E-03}&5.5241E-03&1.1178E-03&1\\
$f_{11}$ & 4.8161E+00&2.6975E+01&6.3374E-01&2&8.7394E+00&5.1116E+01&6.9625E-01&4\\
$f_{12}$ & \bf{1.0936E-07}&9.0059E-07&3.1753E-08&1&2.1104E+00&2.6246E+01&3.1908E-08&4\\
$f_{13}$ & 1.4290E-06&1.3039E-05&3.9950E-07&2&\bf{1.0141E-06}&1.0198E-05&2.2801E-07&1\\ 
Avg. & & & &\bf{2.17}& & & &\bf{2.17}\\ \hline
		\end{tabular} 
	\end{center}
\end{table*}

\begin{table*}
	\caption{Simulation Results and Comparisons for Category III}
	\small
	\begin{center}
		\begin{tabular}{l|llll|llll}
			\hline
			\multirow{2}*{Function} & \multicolumn{4}{c|}{CRO\_G} & \multicolumn{4}{c}{CRO\_C} \\ \cline{2-9}
			 & Mean & Std. Div. & Best & Rank & Mean & Std. Div. & Best & Rank\\ \hline
$f_{14}$ & 3.3089E+00&2.8636E+01&9.9800E-01&3&\bf{1.1893E+00}&6.5107E+00&9.9800E-01&1\\
$f_{15}$ & \bf{5.9906E-04}&1.4234E-03&3.5413E-04&1&6.1532E-04&1.3869E-03&3.2870E-04&2\\
$f_{16}$ & \bf{-1.0305E+00}&2.1115E-02&-1.0316E+00&1&-1.0305E+00&2.1463E-02&-1.0316E+00&2\\
$f_{17}$ & 3.9795E-01&9.2041E-04&3.9789E-01&2&\bf{3.9795E-01}&6.8579E-04&3.9789E-01&1\\
$f_{18}$ & \bf{3.0009E+00}&1.7177E-02&3.0000E+00&1&3.0013E+00&1.9859E-02&3.0000E+00&2\\
$f_{19}$ & \bf{-3.8615E+00}&1.1619E-02&-3.8628E+00&1&-3.8612E+00&1.4214E-02&-3.8628E+00&2\\
$f_{20}$ & \bf{-3.3125E+00}&6.7953E-02&-3.3217E+00&1&-3.3107E+00&7.8734E-02&-3.3214E+00&2\\
$f_{21}$ & \bf{-1.0126E+01}&2.3876E-01&-1.0153E+01&1&-1.0117E+01&2.6175E-01&-1.0151E+01&2\\
$f_{22}$ & -1.0308E+01&6.6218E+00&-1.0402E+01&2&\bf{-1.0350E+01}&6.7161E-01&-1.0401E+01&1\\
$f_{23}$ & -1.0076E+01&1.5155E+01&-1.0536E+01&4&-1.0269E+01&1.0470E+01&-1.0536E+01&3\\ 
Avg. & & & &\bf{1.70}& & & &1.80\\ \hline
			\multirow{2}*{Function} & \multicolumn{4}{c|}{CRO\_E} & \multicolumn{4}{c}{CRO\_R} \\ \cline{2-9}
			 & Mean & Std. Div. & Best & Rank & Mean & Std. Div. & Best & Rank\\ \hline
$f_{14}$ & 2.2217E+00&2.0650E+01&9.9800E-01&2&3.7510E+00&3.6642E+01&9.9800E-01&4\\
$f_{15}$ & 6.5008E-04&1.0962E-03&3.6644E-04&4&6.4918E-04&1.4605E-03&3.2389E-04&3\\
$f_{16}$ & -1.0275E+00&1.2423E-01&-1.0316E+00&3&-1.0183E+00&4.3638E-01&-1.0316E+00&4\\
$f_{17}$ & 3.9833E-01&6.0664E-03&3.9789E-01&4&3.9802E-01&1.7142E-03&3.9789E-01&3\\
$f_{18}$ & 3.0043E+00&6.2059E-02&3.0000E+00&4&3.0014E+00&2.0636E-02&3.0000E+00&3\\
$f_{19}$ & -3.8606E+00&2.5016E-02&-3.8627E+00&4&-3.8606E+00&2.1030E-02&-3.8627E+00&3\\
$f_{20}$ & -3.3086E+00&9.0811E-02&-3.3218E+00&3&-3.3073E+00&9.8572E-02&-3.3215E+00&4\\
$f_{21}$ & -9.9484E+00&2.2520E+00&-1.0141E+01&4&-1.0057E+01&5.0092E+00&-1.0153E+01&3\\
$f_{22}$ & -1.0045E+01&7.5628E+00&-1.0383E+01&4&-1.0188E+01&8.9541E+00&-1.0401E+01&3\\
$f_{23}$ & -1.0305E+01&2.3769E+00&-1.0531E+01&2&\bf{-1.0465E+01}&8.0594E-01&-1.0534E+01&1\\ 
Avg. & & & &3.40& & & &3.10\\ \hline
		\end{tabular} 
	\end{center}
\end{table*}

From these tables we can see the Cauchy distribution performs best in Category I as its averaged rank is the highest. Recall that Category I is uni-modal and thus algorithms performing local search are good enough for these functions. Comparing Figs. 1, 2, 3, and 5, the Cauchy distribution has the most of its probabilities around the mean ($x=0$), which helps CRO perform local search more efficiently for uni-modal problems.

For Category II, the exponential distribution and the Rayleigh distribution perform equally but their advantage over the Gaussian and the Cauchy distributions is not very significant. Since the solution space of problems in Category II is high dimensional with a large number of local optimums, it is reasonable that different distributions will give a best performance on different problems. For problems with relatively less local optimums, the Gaussian and the Cauchy distribution can perform better because these problems are similar with problems in Category I. However, for problems with a large number of local optimums, a flatter distribution can probably perform well since it can maintain the population diversity better. So for this category of problems different distributions can perform well on different problems and there is no significant preference.

The Gaussian distribution performs best in Category III but the Cauchy distribution also performs relatively well in this category. A possible reason to this phenomena is that since the difference between Category II and III is that Category III has lower dimensions, which reduce the number of potential local optimums, the exponential and the Rayleigh distributions lose their advantage on maintaining population diversity. However, although the number of local optimums is smaller, it is still easy for the Cauchy distribution to get stuck due to the shape of its PDF. The Gaussian distribution with shallower bell shape can maintain a better population diversity than the Cauchy while performing local search more efficiently than the other two distributions.

The average computational time consumed in a simulation run of these four CRO variants are listed in Table VI for reference. Generally the computational speed of the Gaussian distribution is the fastest, but all computational times are comparable. To summarize, the Cauchy distribution is suitable for solving uni-modal optimization problems. The Gaussian distribution is suitable for solving low-dimensional multi-modal problem. For high-dimensional multi-modal problems, the exponential and the Rayleigh distribution generally perform better, but for different problems there are different preferences.

\begin{table}
	\caption{Computational Time (s)}
	\small
	\begin{center}
		\begin{tabular}{l|llll}
			\hline
			Function & CRO\_G & CRO\_C & CRO\_E & CRO\_R  \\ \hline
			$f_{1}$ & 0.03281&0.03812&0.04027&0.04674\\
			$f_{2}$ & 0.03346&0.03992&0.04295&0.04835\\
			$f_{3}$ & 0.14172&0.15247&0.15339&0.16701\\
			$f_{4}$ & 0.04294&0.04648&0.04752&0.05661\\
			$f_{5}$ & 0.03841&0.04318&0.0421&0.05111\\
			$f_{6}$ & 0.26363&0.26166&0.26285&0.26445\\
			$f_{7}$ & 0.44546&0.45033&0.45933&0.44639\\
			$f_{8}$ & 0.77906&0.76936&0.80069&0.76243\\
			$f_{9}$ & 0.61241&0.59982&0.61947&0.61142\\
			$f_{10}$ & 0.40227&0.39936&0.40278&0.40402\\
			$f_{11}$ & 0.77743&0.79861&0.80926&0.8098\\
			$f_{12}$ & 0.35557&0.37177&0.3812&0.38371\\
			$f_{13}$ & 0.37318&0.38145&0.38168&0.39597\\
			$f_{14}$ & 0.03674&0.03799&0.03733&0.03802\\
			$f_{15}$ & 0.0631&0.07312&0.07215&0.08616\\
			$f_{16}$ & 0.00058&0.0006&0.00059&0.00066\\
			$f_{17}$ & 0.00131&0.00144&0.00141&0.0017\\
			$f_{18}$ & 0.00174&0.00212&0.00206&0.0026\\
			$f_{19}$ & 0.00211&0.00233&0.00229&0.00245\\
			$f_{20}$ & 0.00444&0.00461&0.00441&0.00485\\
			$f_{21}$ & 0.00202&0.00234&0.00233&0.00297\\
			$f_{22}$ & 0.00221&0.00266&0.00245&0.00312\\
			$f_{23}$ & 0.00248&0.00281&0.0028&0.00336\\ \hline
		\end{tabular} 
	\end{center}
\end{table}
\section{Conclusion and Future Work}
In this paper, we compare four kinds of distribution functions, namely the Gaussian distribution, the Cauchy distribution, the exponential distribution, and the modified Rayleigh distribution, as the perturbation function for CRO. Since these distributions have different characteristics, they may be suitable to solve different kinds of problems with different solution space characteristics. We integrate these four distribution functions into CRO. The four CRO variants are evaluated with 23 benchmark functions, divided into three categories. The simulation results show that different categories of problems have different preference on the perturbation function. The Cauchy distribution fits Category I functions best and the Gaussian distribution fits Category III functions best, while the exponential and the modified Rayleigh distribution perform similarly for the Category II functions and outperform the other two distributions. Our study gives guidelines to design CRO for different types of optimization problems.

In the future we will conduct a systematic analysis on the impact of different parameters on perturbation function selection. We will also perform an analysis on the convergence speed of the four different distribution functions. Other distribution functions, e.g. the Levy distribution, will also be implemented on CRO to have a more general study of the overall performance.

\section*{Acknowledgement}

This work is supported in part by the Strategic Research Theme of Information Technology of The University of Hong Kong. A.Y.S. Lam is also supported in part by the Croucher Foundation Research Fellowship.

\end{document}